\pdfoutput=1
\documentclass[namedate,unnumsec,webpdf,modern,large]{oup-authoring-template}

\graphicspath{{Fig/}}
\newcommand{\societylogo}{}

\theoremstyle{thmstyleone}%
\theoremstyle{thmstyletwo}%
\theoremstyle{thmstylethree}%

\usepackage{graphicx}        
\usepackage{multirow}        
\usepackage[table]{xcolor}   
\usepackage{makecell}        
\usepackage{amsmath}         
\usepackage{pifont}
\usepackage[linesnumbered,ruled,vlined,algo2e]{algorithm2e}

\def\method{scOPD}

\definecolor{lightpink}{HTML}{FFF2F2}

\definecolor{linkblue}{HTML}{2221FC}
\definecolor{lightyellow}{HTML}{FEFDE9}
\definecolor{lightper}{HTML}{F2F2FF}
\definecolor{sum}{HTML}{FFAF59}
\definecolor{arg}{HTML}{4497B8}
\definecolor{eval}{HTML}{778BD2}
\definecolor{red}{HTML}{FF0000}
\definecolor{green}{HTML}{689157}
\definecolor{down}{rgb}{0.8, 0.0, 0.0}
\definecolor{up}{HTML}{4E9C71}
\definecolor{tabletitle}{HTML}{DAD9E4}
\definecolor{mark}{HTML}{000000}
\definecolor{linkgreen}{HTML}{3C7E7F}

\hypersetup{
    colorlinks=true,
    linkcolor=linkblue,
    citecolor=linkblue,
    urlcolor=linkblue
}

\begin{document}
\flushbottom

\journaltitle{Bioinformatics}
\DOI{DOI added during production}
\copyrightyear{YEAR}
\pubyear{YEAR}
\vol{XX}
\issue{x}
\access{Published: Date added during production}
\appnotes{Paper}

\firstpage{1}


\title[scOPD]
{scOPD: Routed On-Policy Self-Distillation for LLM-Based Single-Cell Annotation}









\author[1,2]{Zhiyin Yu}                                    \author[3]{Yuchen Mou}                                     \author[2]{Songtao Huang}                                  \author[2]{Shuaiyu Zhang}                                  \author[4]{Xing Wei}                                       \author[5]{Chunchun Chen}                                  \author[4]{Wei Ye}                                         \author[6]{Xiao Luo}                                       \author[1,$\ast$]{Zhonghai Wu}                            \author[2]{Xiangchao Yan} 
\author[2]{Lei Bai} 
\author[2,$\ast$]{Shiyang Feng}
\author[2]{Bo Zhang}                                       
    
\address[1]{                                          \orgdiv{School of Software and Microelectronics},
\orgname{Peking University},
\orgaddress{Beijing 100871, China}}

\address[2]{
\orgname{Shanghai Artificial Intelligence Laboratory},
\orgaddress{Shanghai 200232, China}}                                                                                  \address[3]{                                               \orgname{Independent Researcher},                       
\orgaddress{Beijing 100000, China}}

\address[4]{
\orgdiv{College of Electronic and Information Engineering},
\orgname{Tongji University},                               \orgaddress{Shanghai 200092, China}}                                                                                
\address[5]{                                               \orgdiv{Shanghai Research Institute for Intelligent Autonomous Systems},                                       \orgname{Tongji University},                               \orgaddress{Shanghai 200092, China}}
 
\address[6]{                                           
\orgname{University of Wisconsin--Madison},
\orgaddress{Madison, WI 53706, USA}}

\corresp[$\ast$]{Corresponding authors. 
\href{fengshiyang@pjlab.org.cn}{fengshiyang@pjlab.org.cn}; \href{mailto:wuzh@pku.edu.cn}{wuzh@pku.edu.cn}}




\received{Date}{0}{Year}
\revised{Date}{0}{Year}
\accepted{Date}{0}{Year}





\abstract{\subheading{Motivation:} Large language models (LLMs) have shown promise for single-cell annotation, but existing training methods mainly rely on scalar outcome rewards that provide limited supervision for intermediate biological reasoning. Batch-level annotation is particularly challenging because it requires jointly resolving local cell identities and a global one-to-one matching constraint, such that one local error can affect the entire assignment. Moreover, a single inference pass provides insufficient evidence for locating uncertain predictions. These limitations call for fine-grained reasoning supervision during training and uncertainty-aware refinement during inference.


\subheading{Results:}
To address these challenges, we propose scOPD, an on-policy self-distillation framework for LLM-based single-cell annotation. After reasoning fine-tuning, scOPD adaptively routes qualified rollouts to reward-based optimization, while erroneous rollouts receive privileged token-level supervision from successful sibling trajectories. During inference, rollout
disagreement triggers constraint-aware pairwise refinement of uncertain
assignments. Across CellPuzzles and four out-of-domain disease datasets, scOPD consistently improves the performance of diverse LLM backbones, achieving average cell-level and batch-level accuracies of 81.71\% and 44.49\% with Qwen3-8B.

\subheading{Availability and Implementation:} Code and data are available at \url{https://github.com/YuZhiyin/scOPD}.

} 







\maketitle


\section{Introduction}

Recent advances in single-cell RNA sequencing (scRNA-seq) have enabled transcriptome-wide profiling at single-cell resolution, revealing the cellular heterogeneity that is obscured by bulk measurements~\citep{kiselev2019challenges,rosenberg2018single}. A fundamental step in scRNA-seq analysis is cell-type annotation, which assigns biologically meaningful identities to individual cells or cell clusters and provides the basis for studying
tissue composition and disease-associated cellular changes~\citep{10.1093/bioinformatics/btae128}. Traditional workflows typically follow a cluster-then-annotate paradigm, where cells are grouped based on their expression profiles and subsequently annotated by matching cluster-specific genes with canonical markers from the literature or curated databases~\citep{10.1093/bioinformatics/btz592,10.1093/bioinformatics/btab840}. Although widely used, this procedure is
labor-intensive and dependent on expert knowledge, and its results can vary with clustering resolution, marker availability, and the ambiguity
of marker expression~\citep{abdelaal2019comparison,yang2022scbert}. Correlation-based methods provide an alternative by transferring labels from annotated datasets to query cells according to similarities between their expression profiles~\citep{kiselev2018scmap,aran2019reference}. However, their performance depends on the quality and coverage of reference datasets~\citep{abdelaal2019comparison}. These limitations motivate more scalable annotation methods that can learn robust cellular representations across heterogeneous datasets.


To achieve scalable and automated annotation, researchers have developed supervised learning methods. For example, scClassify uses an ensemble of weighted k-nearest-neighbor classifiers, while SingleCellNet employs a random forest classifier~\citep{lin2020scclassify,tan2019singlecellnet}. Deep-learning approaches further extend this paradigm: ACTINN uses a multilayer neural network to predict cell types, whereas JIND learns a discriminative latent representation and aligns query cells to the reference space to mitigate batch effects~\citep{10.1093/bioinformatics/btz592,10.1093/bioinformatics/btac140}.
More recently, single-cell foundation models such as scBERT, Geneformer, scGPT, and scFoundation have been pretrained on large-scale transcriptomic data to learn transferable gene and cell representations~\citep{yang2022scbert,theodoris2023transfer,cui2024scgpt,hao2024large}. However, their zero-shot annotation performance remains limited, and adaptation to specific annotation tasks often requires labeled data and task-specific fine-tuning~\citep{kedzierska2025zero}.

Recently, large language models (LLMs) such as GPT-4 have demonstrated
promising zero-shot capabilities for cell-type annotation
~\citep{hou2024assessing}. To bridge
the modality gap between gene-expression profiles and natural language, existing studies transform biological profiles into LLM-compatible representations, such as cell sentences~\citep{Rizvi2025.04.14.648850} and semantic gene embeddings~\citep{chen2025simple}. Subsequent approaches improve LLM-based single-cell analysis through domain-specific adaptation~\citep{schaefer2025multimodal}, retrieval-augmented generation~\citep{yu-etal-2025-scrag,mao2025scagentuniversalsinglecellannotation}, and multi-agent collaboration~\citep{xiao2024cellagent,xie2025cassia,yang2026large}. Beyond annotating cells independently, Cell-o1 formulates batch-level annotation as a constrained reasoning problem in which each candidate cell type can be assigned only
once, and combines reasoning fine-tuning with reinforcement learning
~\citep{10.1093/bioinformatics/btag208}. These advances demonstrate the potential of LLMs to move beyond traditional classification by interpreting biological evidence and generating explicit reasoning for cell-type assignments.

\begin{figure}[t]
    \centering
    \includegraphics[width=\columnwidth]{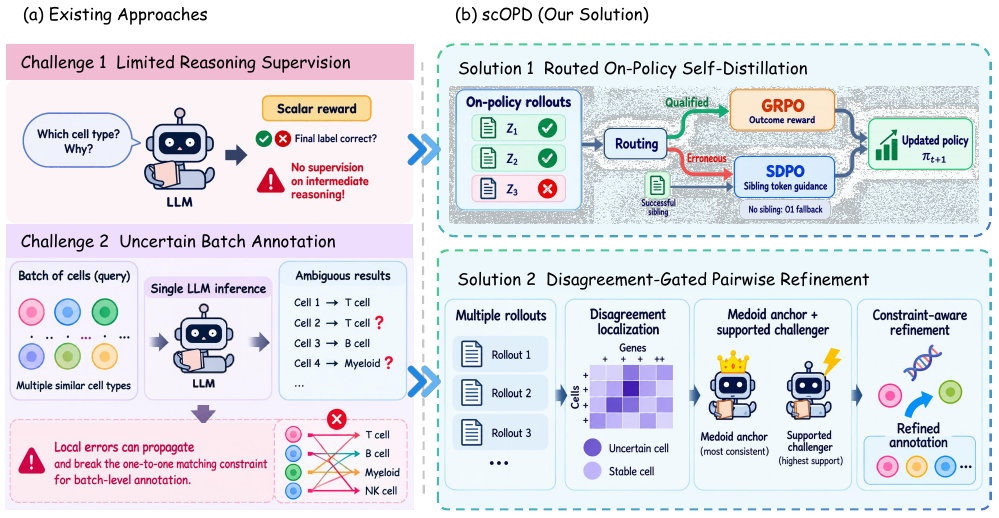}
    \caption{Comparison between (a) existing LLM-based single-cell annotation approaches and (b) \method{}.}
    \vspace{-6mm}
    \label{fig:introduction}
\end{figure}

Despite recent advances in LLM-based single-cell annotation, the following limitations remain to be addressed: (i) Fine-grained supervision for biological reasoning is difficult to obtain. Existing reinforcement learning approaches mainly optimize scalar rewards computed from final assignments, providing only coarse response-level feedback without indicating which token-level reasoning steps are reliable or erroneous~\citep{hübotter2026reinforcementlearningselfdistillation,li2026unifyinggrouprelativeselfdistillationpolicy}. (ii) Identifying and refining uncertain assignments remains challenging. Ambiguity between similar cell types can cause local errors to propagate across the entire assignment, while a single inference pass provides limited evidence for identifying uncertain cells~\citep{xie2025cassia}. Therefore, an effective framework is needed to exploit complementary information across multiple trajectories for fine-grained training supervision and uncertainty-aware refinement.

To address these limitations, we propose \underline{s}ingle-\underline{c}ell \underline{O}n-\underline{P}olicy Self-\underline{D}istillation (\method{}), a framework that exploits complementary information across on-policy rollouts during both training and inference (see Figure~\ref{fig:introduction}). Specifically, we first perform reasoning fine-tuning on OpenAI O1-distilled trajectories to initialize biologically grounded reasoning. Then, we introduce routed on-policy self-distillation: qualified successful rollouts are optimized through outcome-based reinforcement learning, whereas erroneous rollouts receive dense token-level supervision from successful siblings. When no qualified sibling is available, an offline OpenAI-O1 trajectory provides fallback supervision, and entropy-aware weighting emphasizes reliable teacher predictions. During inference, we further introduce disagreement-gated pairwise refinement, which identifies uncertain cells from cross-rollout disagreement and revises them using a medoid anchor and a supported challenger. In this way, \method{} transforms rollout consensus into fine-grained reasoning supervision and rollout disagreement into uncertainty-aware refinement.

We conduct extensive experiments to compare \method{} with general-purpose LLMs, reasoning-fine-tuned LLMs, and domain-specific annotation methods on the in-domain CellPuzzles benchmark and four out-of-domain disease-specific datasets. Across four backbone models, \method{} consistently improves single-cell annotation performance and achieves substantial gains in batch-level accuracy. Further analyses show that the proposed routing mechanism automatically forms an adaptive training curriculum, while rollout disagreement provides an effective signal for uncertainty-aware refinement. These results demonstrate the effectiveness, generalizability, and robustness of \method{} for LLM-based single-cell annotation.

\begin{figure*}[t]
    \centering
    \includegraphics[width=\linewidth]{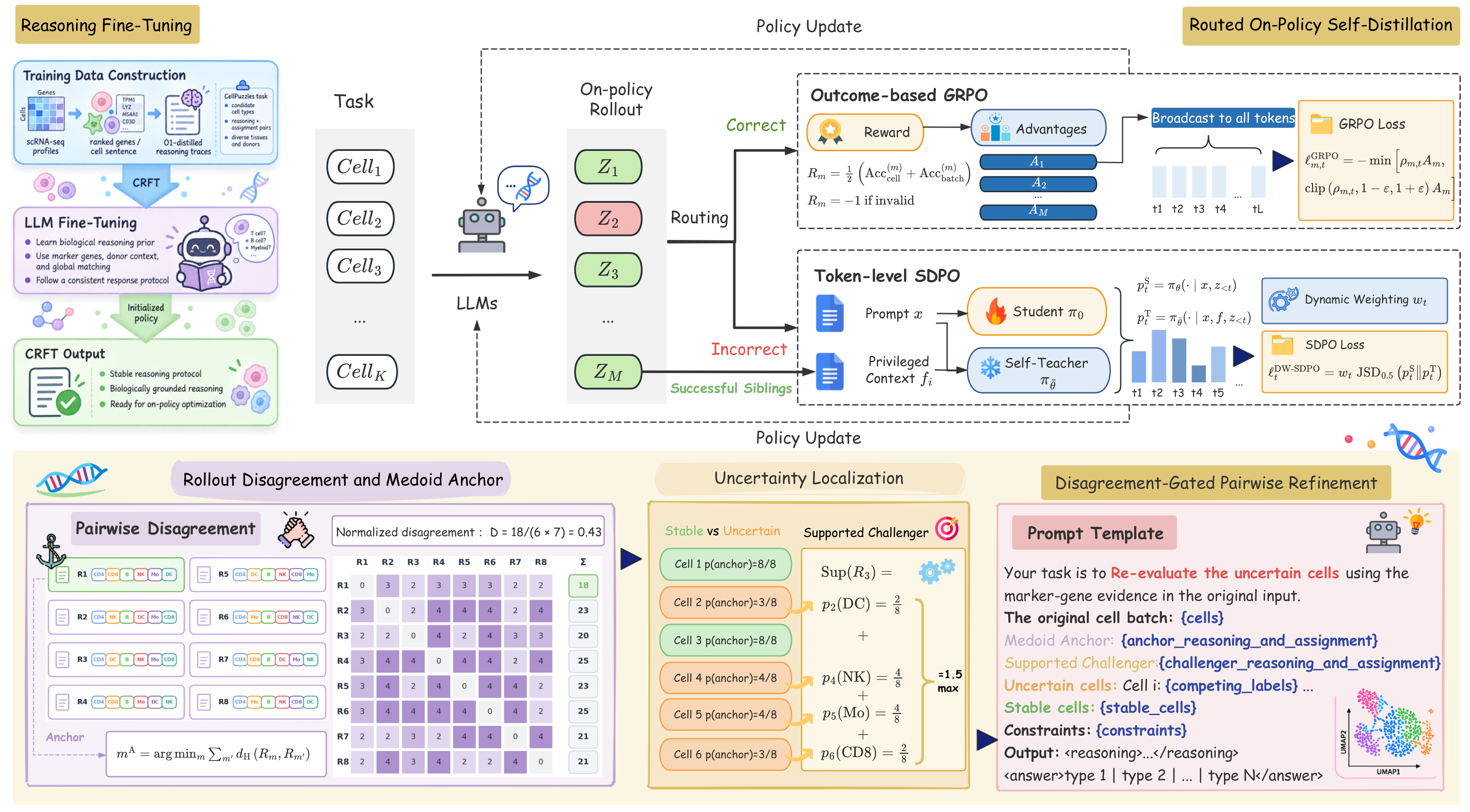}
    \vspace{-16pt}
    \caption{\textbf{Overview of \method{}.} The framework consists of three stages: (1) Cold-Start Reasoning Fine-Tuning, which initializes biologically grounded reasoning from OpenAI o1-distilled trajectories; (2) Routed On-Policy Self-Distillation, which routes qualified rollouts to outcome-based GRPO and erroneous rollouts to entropy-weighted SDPO using successful siblings as privileged context; and (3)Disagreement-Gated Pairwise Refinement, which uses rollout disagreement to refine uncertain assignments.}
    \vspace{-5mm}
    \label{fig:framework}
\end{figure*}

\section{Materials and methods}


\subsection{Overview of \method{}}
As illustrated in Figure~\ref{fig:framework}, our \method{} framework is a multi-stage process that enhances both cell-level and batch-level annotation accuracy. First, the base model performs reasoning fine-tuning by distilling OpenAI o1-reasoning trajectories to form an initial biological reasoning prior. Second, during training, we sample multiple rollouts from the evolving policy and adaptively route them based on rollout correctness to either outcome-based reward optimization or token-level self-distillation. Finally, during inference, disagreement-gated pairwise refinement identifies uncertain cells and revises them using a medoid anchor and a supported challenger while preserving stable assignments.



\subsection{Cold-Start Reasoning Fine-Tuning}

General-purpose LLMs show limited zero-shot performance on
single-cell annotation and often fail to follow the required structured output protocol~\citep{10.1093/bioinformatics/btag208}. To address these limitations, we introduce Cold-Start Reasoning Fine-Tuning (CRFT) to establish an initial reasoning ability and a consistent output format before on-policy optimization.


Let
$\mathcal{D}_{\mathrm{R}}
=\{(x_k,r_k^{*},\mathbf{y}_k^{*})\}_{k=1}^{N_{\mathrm{R}}}$
denote the reasoning fine-tuning dataset containing
$N_{\mathrm{R}}$ training examples.
For the $k$-th example, $x_k$ is an annotation task involving
a batch of $n_k$ cells, $r_k^{*}$ is an OpenAI o1-distilled
reasoning trajectory, and
$\mathbf{y}_k^{*}=(y_{k,1}^{*},\ldots,y_{k,n_k}^{*})$
is the corresponding ground-truth cell-type assignment,
where $y_{k,i}^{*}$ denotes the label of cell $i$ in task $k$.
We construct each target response as
\begin{equation}
    z_k^{*}
    =
    \underbrace{
    \texttt{<reasoning>}
    \;r_k^{*}\;
    \texttt{</reasoning>}
    }_{\text{explicit biological reasoning}}
    \oplus
    \underbrace{
    \texttt{<answer>}
    \;\mathbf{y}_k^{*}\;
    \texttt{</answer>}
    }_{\text{cell-type assignment}},
    \label{eq:crft_target}
\end{equation}
where $\oplus$ denotes sequence concatenation. The instruction requires the model to ground its assignments in marker genes, cellular and donor context, and the global matching constraint.
Specifically, let $\mathcal{T}_k$ denote the set of $n_k$
candidate cell types provided with $x_k$. Under the task protocol considered here, each candidate type
must be assigned to exactly one cell.

Given the tokenized target response
$z_k^{*}=(z_{k,1}^{*},\ldots,z_{k,T_k}^{*})$, where $T_k$ denotes the number of assistant tokens, CRFT minimizes the completion-only negative log-likelihood:
\begin{equation}
    \mathcal{L}_{\mathrm{CRFT}}
    =
    -\frac{1}{\sum_{k=1}^{N_{\mathrm{R}}}T_k}
    \sum_{k=1}^{N_{\mathrm{R}}}
    \sum_{t=1}^{T_k}
    \log
    \pi_{\theta}
    \left(
    z_{k,t}^{*}
    \mid
    x_k,z_{k,<t}^{*}
    \right),
    \label{eq:crft_objective}
\end{equation}
where $\pi_{\theta}$ denotes the LLM policy, and $z_{k,<t}^{*}$ denotes the ground-truth
response prefix preceding token $t$. After optimization, the resulting policy $\pi_{\theta_{\mathrm{CRFT}}}$ initializes the training stage. CRFT establishes a stable protocol, which are refined through on-policy training with successful trajectories from the evolving policy.


\subsection{Routed On-Policy Self-Distillation}

Although Cold-Start Reasoning Fine-Tuning provides an initial reasoning capability, it relies on static offline trajectories and cannot adapt its supervision to the evolving policy~\citep{agarwal2024onpolicy}. To address this limitation, we introduce Routed On-Policy Self-Distillation (ROPSD), which unifies Group-Relative Policy Optimization (GRPO) and Self-Distillation Policy Optimization (SDPO) through rollout-level routing. Before presenting the routing mechanism, we first introduce the two complementary optimization signals.

\noindent \textbf{\textit{Group-Relative Policy Optimization.}} 
Given an annotation task $x_k$,
GRPO~\citep{shao2024deepseekmathpushinglimitsmathematical}
samples a group of $M$ responses $\{z_k^{(m)}\}_{m=1}^{M}$
from the LLM $\pi_{\theta}$.
Let $\hat{\mathbf{y}}_k^{(m)}
=(\hat{y}_{k,1}^{(m)},\ldots,\hat{y}_{k,n_k}^{(m)})$
denote the predicted cell-type assignment for task $x_k$.
Cell-level accuracy measures the fraction of correctly annotated
cells, whereas batch-level exact match requires all cell labels
in the task to be correct.
For structurally valid responses, we define the dense reward
as their average:
\begin{equation}
R_k^{(m)} = \frac{1}{2}
\left[
\frac{1}{n_k}\sum_{i=1}^{n_k}
\mathbb{I}\left(\hat{y}_{k,i}^{(m)}=y_{k,i}^{*}\right)
+
\mathbb{I}\left(\hat{\mathbf{y}}_k^{(m)}=\mathbf{y}_k^{*}\right)
\right],
\label{eq:dense_reward}
\end{equation}
where $\mathbb{I}(\cdot)$ denotes the indicator function.
Responses with an invalid format, an incorrect number of labels,
or duplicated assignments receive $R_k^{(m)}=-1$.
We estimate the advantage of each response by normalizing
its reward within the group:
\begin{equation}
A_k^{(m)}
    =
    \frac{
    R_k^{(m)}
    -
    \operatorname{mean}
    \left(
    \{R_k^{(1)},\ldots,R_k^{(M)}\}
    \right)
    }{
    \operatorname{std}
    \left(
    \{R_k^{(1)},\ldots,R_k^{(M)}\}
    \right)
    +\epsilon
    }.
\end{equation}
The policy is optimized using the clipped objective:
\begin{equation}
    \ell^{\mathrm{GRPO}}
    =
    -\min\left\{
    \rho_{k,m,t}A_k^{(m)},
    \operatorname{clip}
    \left(\rho_{k,m,t},1-\varepsilon,1+\varepsilon\right)
    A_k^{(m)}
    \right\}.
    \label{eq:grpo_objective}
\end{equation}
where $\rho_{k,m,t}$ is the probability ratio between the current and rollout policies at token $t$. GRPO aligns the policy with both cell- and batch-level correctness, but assigns the same outcome-derived advantage to every token in a response, providing only coarse supervision for intermediate reasoning.

\noindent \textbf{\textit{Self-Distillation Policy Optimization.}}
SDPO~\citep{hübotter2026reinforcementlearningselfdistillation} complements GRPO by providing token-level supervision from a privileged self-teacher. For each on-policy response $z_k^{(m)}$, the student and teacher distributions at token position $t$ are defined as
$ p_{k,m,t}^{\mathrm{S}}
    =
    \pi_{\theta}
    \left(
    \cdot
    \mid
    x_k,z_{k,<t}^{(m)}
    \right)$ and $p_{k,m,t}^{\mathrm{T}}
    =
    \pi_{\bar{\theta}}
    \left(
    \cdot
    \mid
    x_k,f_k^{(m)},z_{k,<t}^{(m)}
    \right)$,
where $f_k^{(m)}$ denotes the privileged context available to the teacher, and $\bar{\theta}$ denotes the exponential-moving-average teacher parameters updated by
$\bar{\theta}\leftarrow(1-\eta)\bar{\theta}+\eta\theta$, with $\eta\in(0,1]$ is the EMA update coefficient. The teacher re-scores the student's trajectory under the privileged context. Let
$p_{k,m,t}^{\mathrm{mix}}
=(1-\alpha)p_{k,m,t}^{\mathrm{S}}
+\alpha p_{k,m,t}^{\mathrm{T}}$.
The token-level SDPO loss is defined using the generalized Jensen--Shannon divergence:
\begin{equation}
    \ell^{\mathrm{SDPO}}
    =
    (1-\alpha)
    D_{\mathrm{KL}}
    \left(
    p_{k,m,t}^{\mathrm{S}}
    \Vert
    p_{k,m,t}^{\mathrm{mix}}
    \right)
    +
    \alpha
    D_{\mathrm{KL}}
    \left(
    p_{k,m,t}^{\mathrm{T}}
    \Vert
    p_{k,m,t}^{\mathrm{mix}}
    \right).
    \label{eq:sdpo_loss}
\end{equation}
We set $\alpha=0.5$, yielding the symmetric Jensen--Shannon divergence~\citep{10.1109/18.61115}. Unlike GRPO, which assigns one outcome-derived advantage to the entire response, SDPO supervises the next-token distributions along the student's on-policy trajectory and provides guidance for its reasoning.

\noindent \textbf{\textit{Hierarchical Rollout Routing.}}
Applying SDPO to all rollouts may introduce unnecessary supervision on correct trajectories~\citep{li2026unifyinggrouprelativeselfdistillationpolicy}. Hence, we route each rollout according to its correctness and the availability of privileged information. A rollout is considered \emph{qualified} if its batch assignment is correct, and its response follows the required format. For an unqualified rollout, the privileged context is selected hierarchically as 
\begin{equation} f_k^{(m)} = \begin{cases} z_k^{(m')}, & \text{if a qualified sibling } m'\neq m \text{ exists},\\ z_k^{\mathrm{O1}}, & \text{if no qualified sibling exists and O1 is available},\\ \varnothing, & \text{otherwise}. \end{cases} \label{eq:scopd_privileged_context} \end{equation} 

Accordingly, qualified rollouts are optimized with GRPO, while unqualified rollouts with a qualified sibling are optimized with sibling SDPO. If no qualified sibling exists, an OpenAI-O1 trajectory provides fallback supervision for an incorrect rollout.
All remaining rollouts fall back to GRPO. We use $s_k^{(m)}$ and $g_k^{(m)}$ as the indicators of the SDPO and GRPO branches.

\noindent \textbf{\textit{Entropy-Aware Dynamic Weighting.}}
To stabilize training under varying teacher confidence, we weight each SDPO token according to the entropy of the teacher distribution:
\begin{equation}
    H_{k,m,t}^{\mathrm{T}}
    =
    -
    \sum_{v\in\mathcal{W}}
    p_{k,m,t}^{\mathrm{T}}(v)
    \log p_{k,m,t}^{\mathrm{T}}(v),
    \label{eq:scopd_teacher_entropy}
\end{equation}
where $\mathcal{W}$ denotes the vocabulary. We convert the entropy into
an unnormalized confidence weight
$\widetilde{w}_{k,m,t}
=\exp\left(-\beta H_{k,m,t}^{\mathrm{T}}\right)$,
where $\beta>0$ controls the sensitivity to teacher uncertainty. The
weight is normalized over all response tokens routed to SDPO:
\begin{equation}
    w_{k,m,t}
    =
    \frac{
    \widetilde{w}_{k,m,t}
    }{
    \displaystyle
    \frac{1}{|\Omega_{\mathrm{S}}|}
    \sum_{(k',m',t')\in\Omega_{\mathrm{S}}}
    \widetilde{w}_{k',m',t'}
    },
    \label{eq:scopd_dynamic_weight}
\end{equation}
where $\Omega_{\mathrm{S}}$ denotes the set of valid tokens routed to SDPO. Thus, confident low-entropy predictions receive larger weights, while uncertain predictions are downweighted. The dynamically weighted SDPO loss is then
$\ell^{\mathrm{DW\text{-}SDPO}}
    =
    w_{k,m,t}
    \ell_{k,m,t}^{\mathrm{SDPO}}.
    \label{eq:scopd_dw_sdpo}$

\noindent \textbf{\textit{Unified Training Objective.}}
Although the rewards of all $M$ rollouts are used to compute the group-relative advantages, these advantages contribute to the loss only for rollouts assigned to the GRPO branch. Rollouts assigned to SDPO are instead optimized directly through the teacher--student divergence. The final routed objective is: 
\begin{equation}
    \mathcal{L}_{\mathrm{ROPSD}}
    =
    \frac{
    \displaystyle
    \sum_{k,m,t}
    \left[
    g_k^{(m)}
    \ell^{\mathrm{GRPO}}
    +
    s_k^{(m)}
    \ell^{\mathrm{DW\text{-}SDPO}}
    \right]
    }{
    \displaystyle
    \sum_{k,m,t}
    \left[
    g_k^{(m)}+s_k^{(m)}
    \right]
    }. 
    \label{eq:scopd_unified_objective}
\end{equation}

The two branches are aggregated according to their routed response tokens. As more rollouts become qualified during training, the optimization gradually shifts from SDPO to GRPO, forming an adaptive curriculum driven by policy quality.

\begin{algorithm2e}[t]
\caption{The \method{} framework}
\label{alg:scopd}

\SetKwInOut{Input}{Input}
\SetKwInOut{Output}{Output}

\Input{
SFT dataset $\mathcal{D}_{\mathrm{R}}$,
ROPSD dataset $\mathcal{D}_{\mathrm{train}}$,
base policy $\pi_{\theta_0}$,
and rollout number $M$
}
\Output{Trained policy $\pi_{\theta}$ and refined predictions}

\tcp{\textcolor{linkgreen}{\textbf{Stage 1: Cold-Start Reasoning Fine-Tuning}}}
Train $\pi_{\theta_0}$ on $\mathcal{D}_{\mathrm{R}}$ using
$\mathcal{L}_{\mathrm{CRFT}}$ by
Eq.~\eqref{eq:crft_objective}\; 

Initialize the policy $\pi_{\theta}$ and EMA teacher
$\pi_{\bar{\theta}}$\;

\tcp{\textcolor{linkgreen}{\textbf{Stage 2: Routed On-Policy Self-Distillation}}}
\ForEach{$x_k\in\mathcal{D}_{\mathrm{train}}$}{
    Sample $M$ rollouts
    $\{z_k^{(m)}\}_{m=1}^{M}$ from $\pi_{\theta}$\;
    
    Compute dense rewards by Eq.~\eqref{eq:dense_reward}, select privileged contexts and route rollouts by
    Eq.~\eqref{eq:scopd_privileged_context}\;
    
    Compute the GRPO loss and the SDPO loss with dynamic weighting by Eqs.~\eqref{eq:grpo_objective},
    \eqref{eq:sdpo_loss}, and~\eqref{eq:scopd_dynamic_weight}\;
    
    Update $\pi_{\theta}$ by
    Eq.~\eqref{eq:scopd_unified_objective}
    and update $\pi_{\bar{\theta}}$\;
}

\tcp{\textcolor{linkgreen}{\textbf{Stage 3: Disagreement-Gated Pairwise Refinement}}}
\ForEach{$x_k\in\mathcal{D}_{\mathrm{test}}$}{
    Generate $M$ rollouts and filter invalid responses\;
    
    Compute rollout disagreement and select the medoid anchor by
    Eqs.~\eqref{eq:dgpr_hamming} and~\eqref{eq:dgpr_anchor}\;
    
    Localize uncertain cells and select a supported challenger by Eq.~\eqref{eq:dgpr_support}\;
    
    Refine uncertain assignments if accepted; otherwise return the anchor\;
}

\end{algorithm2e}

\subsection{Disagreement-Gated Pairwise Refinement}

At inference time, a single inference pass remains sensitive to
ambiguity between similar cell types and provides limited evidence for
identifying uncertain assignments~\citep{wang2023selfconsistency}. To solve this, we introduce Disagreement-Gated Pairwise
Refinement (DGPR), which uses rollout disagreement to identify uncertain cells and performs constraint-aware refinement using a medoid anchor and a supported challenger.

\noindent \textbf{\textit{Rollout Disagreement and Medoid Anchor.}} For an annotation task $x_k$, we sample $M$ independent rollouts
$\{z_k^{(m)}\}_{m=1}^{M}$ from the current LLM. Each rollout
$z_k^{(m)}$ consists of a reasoning trajectory $r_k^{(m)}$ and a
predicted cell-type assignment
$\hat{\mathbf{y}}_k^{(m)}$. 
We retain only rollouts that follow the required response format and assign every candidate type in
$\mathcal{T}_k$ exactly once. Let
$\mathcal{V}_k\subseteq\{1,\ldots,M\}$ denote the indices of these
valid rollouts. For two valid assignments, we define their disagreement as:
\begin{equation}
    d_{\mathrm{H}}
    \left(
    \hat{\mathbf{y}}_k^{(m)},
    \hat{\mathbf{y}}_k^{(m')}
    \right)
    =
    \sum_{i=1}^{n_k}
    \mathbb{I}
    \left[
    \hat{y}_{k,i}^{(m)}
    \neq
    \hat{y}_{k,i}^{(m')}
    \right],
    \label{eq:dgpr_hamming}
\end{equation}
where $m,m'\in\mathcal{V}_k$. We select the medoid rollout as the anchor:
\begin{equation}
    m_k^{\mathrm{A}}
    =
    \arg\min_{m\in\mathcal{V}_k}
    \sum_{m'\in\mathcal{V}_k}
    d_{\mathrm{H}}
    \left(
    \hat{\mathbf{y}}_k^{(m)},
    \hat{\mathbf{y}}_k^{(m')}
    \right),
    \label{eq:dgpr_anchor}
\end{equation} and denote its response and assignment by
$z_k^{\mathrm{A}}=z_k^{(m_k^{\mathrm{A}})}$ and
$\hat{\mathbf{y}}_k^{\mathrm{A}}
=\hat{\mathbf{y}}_k^{(m_k^{\mathrm{A}})}$, respectively. The normalized disagreement associated with the anchor is $D_k =
    \frac{
    \displaystyle
    \sum_{m\in\mathcal{V}_k\setminus\{m_k^{\mathrm{A}}\}}
    d_{\mathrm{H}}
    \left(
    \hat{\mathbf{y}}_k^{\mathrm{A}},
    \hat{\mathbf{y}}_k^{(m)}
    \right)
    }{
    n_k\left(|\mathcal{V}_k|-1\right)
    },
    \label{eq:dgpr_disagreement}$ when $|\mathcal{V}_k|\geq 2$. A larger $D_k$ indicates greater inconsistency among the valid assignments.

\noindent \textbf{\textit{Uncertainty Localization and Supported Challenger.}} For each cell $i$, the support for a candidate type
$\tau\in\mathcal{T}_k$ is defined as the fraction of valid rollouts
that assign $\tau$ to that cell:
\begin{equation}
    p_{k,i}(\tau)
    =
    \frac{1}{|\mathcal{V}_k|}
    \sum_{m\in\mathcal{V}_k}
    \mathbb{I}
    \left[
    \hat{y}_{k,i}^{(m)}=\tau
    \right].
    \label{eq:dgpr_support}
\end{equation}
Given a consensus threshold $\gamma$, a cell is considered
\textbf{stable} if the label assigned by the anchor is also the most
frequent label among the valid rollouts and its support is at least
$\gamma$. All remaining cells are considered \textbf{uncertain}.

We select the challenger from non-anchor rollouts that disagree with the anchor on uncertain cells. The rollout with the highest collective support on uncertain cells is chosen, with ties resolved by larger disagreement. If no eligible challenger exists, DGPR returns the anchor without refinement.

\noindent \textbf{\textit{Constraint-Aware Refinement.}} When refinement is triggered, the annotation task $x_k$, the anchor response $z_k^{\mathrm{A}}$, the
challenger response $z_k^{\mathrm{C}}$, and the stable and uncertain
cell sets are provided to the LLM model. The model is instructed
to re-examine the marker-gene evidence and global matching
relationships for the uncertain cells while preserving the stable
assignments. This produces a refined response
$z_k^{\mathrm{R}}
=(r_k^{\mathrm{R}},\hat{\mathbf{y}}_k^{\mathrm{R}})$. Thus, DGPR permits evidence-supported corrections on uncertain cells
while preserving stable predictions and the global matching constraint.

\subsection{Summarization}
To summarize, the overall workflow of \method{} is presented in Algorithm~\ref{alg:scopd}. First, Cold-Start Reasoning Fine-Tuning distills OpenAI-O1 reasoning trajectories to initialize a stable biological reasoning policy. Next, Routed On-Policy Self-Distillation samples multiple rollouts and adaptively combines outcome-based optimization with token-level self-distillation: qualified rollouts are optimized by GRPO, while erroneous rollouts receive entropy-weighted supervision from successful siblings or an OpenAI-O1 fallback trajectory. Finally, Disagreement-Gated Pairwise Refinement exploits disagreement among rollouts to localize uncertain cells and performs constraint-aware refinement using a medoid anchor and a supported challenger. Overall, \method{} leverages successful siblings for training-time reasoning supervision and rollout disagreement for inference-time refinement, thereby enhancing single-cell annotation.



\begin{table*}[t]
\centering
\caption{\textbf{Performance comparison} of different strategies on single-cell annotation tasks. The \textbf{best} and \underline{second-best} results are highlighted in bold and underlined.}
\label{tab:main_results}
\resizebox{\textwidth}{!}{%
\begin{tabular}{l |  c c c c c c c c c c c c}
\Xhline{1.2pt}
\rowcolor{tabletitle}

& \multicolumn{2}{c}{\textbf{CellPuzzles}}
& \multicolumn{2}{c}{\textbf{Breast Cancer}}
& \multicolumn{2}{c}{\textbf{Melanoma}}
& \multicolumn{2}{c}{\textbf{SLE}}
& \multicolumn{2}{c}{\textbf{Colorectal Cancer}}
& 
&  \\

\rowcolor{tabletitle} \multirow{-2}{*}{\textbf{Methods}}
& Cell & Batch
& Cell & Batch
& Cell & Batch
& Cell & Batch
& Cell & Batch
&\multirow{-2}{*}{\textbf{Avg. Cell}} & \multirow{-2}{*}{\textbf{Avg. Batch}}\\

\Xhline{1.2pt}

\Xhline{1.2pt}

\multicolumn{13}{c}{\textit{General-purpose LLMs (Zero Shot)}} \\ \hline

o1 & $0.6479$ & $0.1900$ &$0.8136$& $0.3200$ & $0.7738$ & $0.1270$&$0.8200$ & $0.3700$ & $0.7845$&$0.3108$ & $0.7680$ &$0.2636$\\ 

\rowcolor{gray!10} o3-mini & $0.5804$ & $0.1352$ &$0.7806$ & $0.2960$ & $0.7341$ &$0.1111$ & $0.7706$ & $0.3200$ &$0.7208$ & $0.1713$ & $0.7173$&$0.2067$  \\

GPT-4o & $0.4248$ & $0.0283$ & $0.5847$ & $0.0560$ & $0.7619$ & $0.1587$ &$0.6576$ & $0.1000$ &$0.5715$ & $0.0239$ & $0.6001$ &$0.0734$ \\

\rowcolor{gray!10} GPT-4o-mini & $0.2581$ & $0.0027$ & $ 0.4171 $ & $0.0000$ & $0.5198$ &$ 0.0000 $ & $0.4129$ & $0.0000$ &$ 0.4549 $ & $0.0000$ & $0.4126$ & $ 0.0005 $ \\


  
Llama3.3-70B-Instruct & $0.6411$ & $0.2969$ & $ 0.4400 $ & $0.0000$ & $0.5337$ &$ 0.0159 $ & $0.3165$ & $0.0000$ &$ 0.4218 $ & $0.0000$ & $0.4706$ & $ 0.0646 $ \\

\rowcolor{gray!10} Qwen3-4B & $0.1059$ & $0.0000$ & $0.1907$ & $0.0000$ & $0.2242$ & $0.0000$ &$0.1247$ & $0.0000$ &$0.1391$ & $0.0000$ & $0.1569$ &$0.0000$ \\

Qwen3-8B & $0.1522$ & $0.0000$ & $0.2100$ & $0.0000$ & $0.2321$ & $0.0000$ &$0.1624$ & $0.0000$ &$0.1772$ & $0.0000$ & $0.1868$ &$0.0000$ \\

\rowcolor{gray!10} Qwen3.5-9B & $0.3670$ & $0.0511$ & $0.5678$ & $0.1280$ & $0.6587$ & $0.2857$ &$0.2932$ & $0.0200$ &$0.5357$ & $0.0757$ & $0.4845$ &$0.1121$ \\

Llama3.1-8B-Instruct & $0.0050$ & $0.0000$ & $0.0075$ & $0.0000$ & $0.0179$ & $0.0000$ &$0.0035$ & $0.0000$ &$0.0147$ & $0.0000$ & $0.0109$ &$0.0000$\\ \hline

\multicolumn{13}{c}{\textit{Instruction-tuned LLMs (Reasoning SFT)}} \\ \hline

\rowcolor{gray!10} Qwen3-4B & $0.6347$ & $0.1671$ & $0.7826$ & $0.2800$ & $0.7778$ & $0.2698$ &$0.6817$ & $0.1300$ &$0.7581$ & $0.2191$ & $0.7270$ &$0.2132$ \\

Qwen3-8B & $0.6393$ & $0.1735$ & $0.7837$ & $0.2560$ & $0.8155$ & $0.3333$ &$0.7368$ & $0.2400$ &$0.7572$ & $0.2231$ & $0.7465$ &$0.2452$ \\

\rowcolor{gray!10} Qwen3.5-9B & $0.6447$ & $0.1653$ & $0.7723$ & $0.2400$ & $0.7401$ & $0.2540$ &$0.7167$ & $0.2000$ &$0.7710$ & $0.2430$ & $0.7290$ &$0.2205$ \\

Llama3.1-8B-Instruct & $0.6229$ & $0.1288$ & $0.7616$ & $0.1840$ & $0.7917$ & $0.2698$ &$0.7229$ & $0.1600$ &$0.7406$ & $0.1873$ & $0.7279$ &$0.1860$ \\ \hline 

\multicolumn{13}{c}{\textit{Domain-specific Methods}} \\ \hline

\rowcolor{gray!10} ChatCell & $0.1059$ & $\text{--}$ & $ 0.1444 $ & $\text{--}$ & $0.1627$ &$\text{--} $ & $0.1132$ & $\text{--}$ &$ 0.1222 $ & $\text{--}$ & $0.1297$ & $ \text{--} $ \\
 
\rowcolor{gray!10}  Cell-o1 & $0.6975$ & $0.2776$ & $\underline{0.8440}$ & $0.4400$ & $\underline{0.8849}$ & $0.6032$ &$0.7624$ & $0.3200$ &$0.7795$ & $0.2869$ & $0.7937$ &$0.3855$ \\

\rowcolor{gray!10} Cell2Sentence & $0.2191$ & $\text{--}$ &$0.7119$ & $\text{--}$ & $0.5794$ &$\text{--}$ & $0.1176$ & $\text{--}$ &$0.5920$ & $\text{--}$ & $0.4440$&$\text{--}$  \\

GenePT & $0.2058$ & $\text{--}$ &$0.1394$& $\text{--}$ & $0.1190$ & $\text{--}$&$0.0918$ & $\text{--}$ & $0.1895$&$\text{--}$ & $0.1491$ &$\text{--}$\\  \hline

\multicolumn{13}{c}{\textit{\method{} (Ours)}} \\ \hline

\rowcolor{gray!10} Qwen3-4B & $0.7129$ & $0.3032$ & $ 0.8297 $ & $0.3680$ & $0.8155$ &$0.3651 $ & $0.7956$ & $\underline{0.4400}$ &$ 0.7789 $ & $0.2430$ & $0.7865$ & $ 0.3439 $ \\
 
Qwen3-8B & $0.7191$ & $0.3114$ & $\textbf{0.8528}$ & $\textbf{0.4880}$ & $\textbf{0.9087}$ & $\textbf{0.6984}$ &$\underline{0.8051}$ & $0.4000$ &$\underline{0.7998}$ & $\underline{0.3267}$ & $\textbf{0.8171}$ &$\textbf{0.4449}$ \\

\rowcolor{gray!10} Qwen3.5-9B & $\underline{0.7285}$ & $\underline{0.3589}$ &$0.8409$ & $\underline{0.4080}$ & $0.8671$ &$\underline{0.4762}$ & $0.7682$ & $0.3500$ &$\textbf{0.8138}$ & $\textbf{0.3665}$ & $\underline{0.8037}$&$\underline{0.3919}$  \\

Llama3.1-8B-Instruct & $\textbf{0.7382}$ & $\textbf{0.3744}$ &$0.8132$& $0.3120$ & $0.8254$ & $0.3492$&$\textbf{0.8510}$ & $\textbf{0.5300}$ & $0.7748$&$0.2629$ & $0.8005$ &$0.3657$\\  

\Xhline{1.2pt}
\end{tabular}
}
\end{table*}

\section{Experimental setup}
\subsection{Datasets}
We conduct experiments on CellPuzzles following Cell-o1~\citep{10.1093/bioinformatics/btag208}, which is constructed from human scRNA-seq datasets curated by CZ CELLxGENE Discover~\citep{10.1093/nar/gkae1142}. Each instance contains $8$--$15$ cells of distinct types from the same donor and batch, represented by ranked highly expressed genes, shared donor metadata, and a shuffled candidate label set with one-to-one assignment. For training, we use the $3{,}912$ accepted OpenAI o1-distilled reasoning trajectories for CRFT and all $6{,}912$ training samples for ROPSD. For in-domain evaluation, we use the CellPuzzles test split with $1{,}095$ samples. We further evaluate out-of-domain generalization on four disease-specific datasets constructed from raw scRNA-seq data following the same CellPuzzles protocol: Breast Cancer ($125$ batches), Melanoma ($63$), Systemic Lupus Erythematosus ($100$), and Colorectal Cancer ($251$).

\subsection{Baseline Methods}


We compare \method{} with three types of baselines as follows. For methods that independently annotate each cell, we report only cell-level accuracy. (1) \textbf{General-purpose LLMs}: We evaluate closed-source LLMs, including GPT-4o-mini, GPT-4o~\citep{openai2024gpt4technicalreport}, o3-mini, and o1~\citep{openai2026openaio1card}, and open-source LLMs, including Llama3.3-70B-Instruct, Qwen3-4B, Qwen3-8B~\citep{yang2025qwen3technicalreport}, Qwen3.5-9B, and Llama3.1-8B-Instruct~\citep{grattafiori2024llama3herdmodels}, all under zero-shot inference. (2) \textbf{Instruction-tuned LLMs}: We perform reasoning fine-tuning on Qwen3-4B, Qwen3-8B, Qwen3.5-9B, and Llama3.1-8B-Instruct. (3) \textbf{Domain-specific Methods}: We compare ChatCell~\citep{fang2024chatcellfacilitatingsinglecellanalysis}, Cell-o1~\citep{10.1093/bioinformatics/btag208}, Cell2Sentence~\citep{Rizvi2025.04.14.648850}, and GenePT~\citep{chen2025simple}. ChatCell employs a T5 encoder--decoder with cell-specific vocabulary adaptation and unified sequence generation, while Cell-o1 fine-tunes LLMs and applies GRPO with sparse outcome rewards. Cell2Sentence fine-tunes a Pythia-410M autoregressive model over gene-expression sentences, while GenePT-s encodes using \texttt{text-embedding-ada-002} and performs $k$-nearest-neighbor classification.

\subsection{Implementation Details}

We conduct training on $8\times140\,\mathrm{GB}$ NVIDIA H200 GPUs. For CRFT, we train for 10 epochs with a learning rate of $5\times10^{-5}$ and a maximum sequence length of 8,192. For ROPSD, we sample $K=8$ on-policy trajectories per prompt and optimize the model for five epochs with a learning rate of $1\times10^{-6}$ and a maximum response length of 4,096. The EMA teacher update rate and entropy-weighting coefficient are set to $0.05$ and $\beta=1$, respectively. During inference, DGPR samples $M=8$ trajectories with a temperature of $0.6$.

\subsection{Evaluation Metrics}

We evaluate single-cell annotation using cell-level and batch-level accuracy. Cell-level accuracy measures the proportion of correctly annotated cells: 
$\mathrm{Acc}_{\mathrm{cell}}
=
\frac{1}{N}\sum_{i=1}^{N}\mathbb{I}(\hat{y}_i=y_i)$. Batch-level accuracy adopts an exact-match criterion, where a batch is correct only if all its cells are correctly annotated:
$\mathrm{Acc}_{\mathrm{batch}}
=
\frac{1}{B}\sum_{b=1}^{B}
\mathbb{I}\left(\forall i\in\mathcal{B}_b,\hat{y}_i=y_i\right)$.



\begin{table*}[t]
\centering
\caption{\textbf{Ablation study} of \method{} with Qwen3-8B. The \textbf{best} and \underline{second-best} results are highlighted in bold and underlined.}
\label{tab:ablation_study}
\resizebox{\textwidth}{!}{%
\begin{tabular}{l | c c c c c c c c c c c c}
\Xhline{1.2pt}

\rowcolor{tabletitle}

& \multicolumn{2}{c}{\textbf{CellPuzzles}}
& \multicolumn{2}{c}{\textbf{Breast Cancer}}
& \multicolumn{2}{c}{\textbf{Melanoma}}
& \multicolumn{2}{c}{\textbf{SLE}}
& \multicolumn{2}{c}{\textbf{Colorectal Cancer}}
& 
&  \\

\rowcolor{tabletitle} \multirow{-2}{*}{\textbf{Variants}}
& Cell & Batch
& Cell & Batch
& Cell & Batch
& Cell & Batch
& Cell & Batch
&\multirow{-2}{*}{\textbf{Avg. Cell}} & \multirow{-2}{*}{\textbf{Avg. Batch}}\\

\Xhline{1.2pt}

\method{}
& $\textbf{0.7191}$ & $\textbf{0.3114}$
& $\textbf{0.8528}$ & $\textbf{0.4880}$
& $0.9087$ & $0.6984$
& $\textbf{0.8051}$ & $\underline{0.4000}$
& $\textbf{0.7998}$ & $\textbf{0.3267}$
& $\textbf{0.8171}$ & $\textbf{0.4449}$ \\

\rowcolor{gray!10} w/o CRFT
& $0.6891$ & $0.2795$
& $0.8086$ & $0.3680$
& $0.7996$ & $0.3492$
& $0.7268$ & $0.3300$
& $0.7610$ & $0.2550$
& $0.7570$ & $0.3163$ \\

w/o ROPSD
& $0.6693$ & $0.2174$
& $0.8118$ & $0.3440$
& $0.8155$ & $0.3492$
& $0.7604$ & $0.2600$
& $0.7780$ & $0.3028$
& $0.7670$ & $0.2947$ \\

\rowcolor{gray!10} w/o Dense Reward
& $0.7021$ & $0.2822$
& $0.8146$ & $0.3600$
& $0.8810$ & $0.6349$
& $0.7529$ & $0.3000$
& $0.7534$ & $0.2351$
& $0.7808$ & $0.3624$ \\

w/o O1 Fallback
& $\underline{0.7129}$ & $0.2932$
& $0.8107$ & $0.3360$
& $\underline{0.9187}$ & $\underline{0.7301}$
& $\underline{0.7990}$ & $0.3400$
& $\underline{0.7914}$ & $\underline{0.3068}$
& $\underline{0.8065}$ & $0.4012$ \\

\rowcolor{gray!10} w/o DGPR
& $0.6976$ & $\underline{0.2977}$
& $\underline{0.8278}$ & $\underline{0.4320}$
& $\textbf{0.9206}$ & $\textbf{0.7302}$
& $0.7964$ & $\textbf{0.4400}$
& $0.7744$ & $0.2749$
& $0.8034$ & $\underline{0.4350}$ \\

\Xhline{1.2pt}
\end{tabular}
}
\end{table*}

\section{Results}

\begin{figure*}[t]
    \centering
    \includegraphics[width=\linewidth]{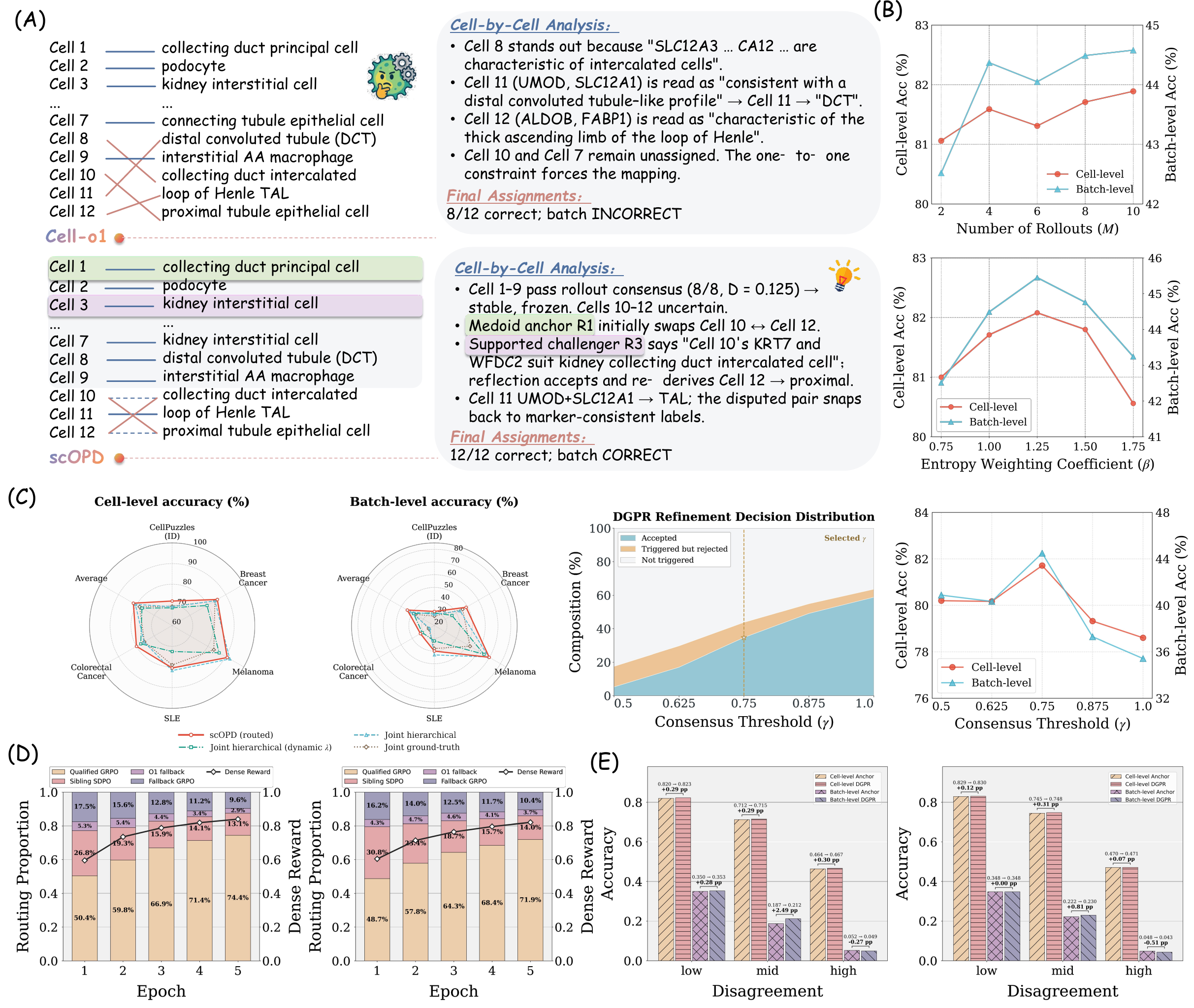}
    \caption{Further analysis of \method{}.
(A) Case study of correcting coupled kidney cell-type assignments by Cell-o1 and \method{}.
(B) Sensitivity analysis of the number of inference rollouts, the entropy-weighting coefficient, and the consensus threshold.
(C) Mechanism analysis comparing training-time strategies.
(D) Adaptive routing analysis across epochs for Qwen3-8B (left) and Qwen3-4B (right).
(E) Disagreement refinement analysis across disagreement levels for Qwen3-8B (left) and Qwen3-4B (right).}
    \vspace{-3mm}
    \label{fig:case}
\end{figure*}

\subsection{Performance Comparison}
We report the main results on the in-domain CellPuzzles benchmark and four out-of-domain disease datasets in Table~\ref{tab:main_results}. We draw the following conclusions.
\textit{\textbf{Firstly,}} \method{} achieves the strongest in-domain performance on CellPuzzles and consistently improves all four backbone models. In particular, \method{} with Llama3.1-8B-Instruct obtains the best cell-level and batch-level accuracies of $0.7382$ and $0.3744$, outperforming the strongest baseline Cell-o1 by $0.0407$ and $0.0968$.
\textit{\textbf{Secondly,}} \method{} demonstrates strong out-of-domain generalization across four unseen disease conditions. With Qwen3-8B, \method{} achieves average out-of-domain accuracies of $0.8416$ and $0.4783$. Moreover, different backbone models attain the best performance on different datasets: Qwen3-8B on Breast Cancer and Melanoma, Llama3.1-8B-Instruct on SLE, and Qwen3.5-9B on Colorectal Cancer. \textit{\textbf{Thirdly,}} \method{} shows a clear advantage on the more challenging batch-level evaluation. Unlike domain-specific baselines such as ChatCell, Cell2Sentence, and GenePT, which annotate cells independently without modeling the one-to-one batch constraint, \method{} jointly reasons over all cells and optimizes both cell- and batch-level correctness. As a result, Qwen3-8B improves from $0.2452$ to $0.4449$ with \method{}.


\subsection{Adaptive Routing Analysis}
To examine the routing dynamics of ROPSD, we visualize the epoch-wise routing proportions and dense rewards of Qwen3-4B and Qwen3-8B in Figure~\ref{fig:case} (D). We have the following observations. \textit{\textbf{Firstly,}} the proportion of qualified GRPO trajectories increases with the dense reward. From the first to the fifth epoch, it rises from $48.7\%$ to $71.9\%$ for Qwen3-4B and from $50.4\%$ to $74.4\%$ for Qwen3-8B. This indicates that more on-policy rollouts gradually achieve both correct assignments and valid reasoning. \textit{\textbf{Secondly,}} as the policy improves, fewer rollouts require token-level self-distillation or fallback handling. The sibling-SDPO proportion decreases from $30.8\%$ to $14.0\%$ for Qwen3-4B and from $26.8\%$ to $13.1\%$ for Qwen3-8B, while the fallback-GRPO proportion also declines from $16.2\%$ to $10.4\%$ and from $17.5\%$ to $9.6\%$, respectively. O1-fallback SDPO remains relatively infrequent, as it is used only when no qualified sibling is available, indicating that successful on-policy siblings are the primary source of privileged information. \textit{\textbf{Thirdly,}} the routing dynamics reveal a learning curriculum: ROPSD relies more on token-level distillation at early stages and shifts toward reward-driven GRPO as the policy improves. This transition is driven by rollout quality rather than a predefined schedule, demonstrating the adaptivity of the routing mechanism.

To further examine the necessity of rollout-level routing, we compare ROPSD with three alternative training-time mechanisms, as shown in Figure~\ref{fig:case} (C). \textit{Joint hierarchical} uses hierarchical privileged contexts with a fixed loss coefficient, \textit{Joint hierarchical (dynamic $\lambda$)} applies source-dependent distillation weights, and \textit{Joint ground-truth} provides ground-truth privileged information to all rollouts. ROPSD achieves the highest average cell-level and batch-level accuracies, outperforming Joint Hierarchical by $1.13$ and $2.74$ percentage points, respectively. These results indicate that uniformly combining GRPO and SDPO can introduce conflicting or unnecessary supervision, whereas routing assigns each rollout the optimization signal most appropriate to its current quality.



\subsection{Disagreement Refinement Analysis} 
Figure~\ref{fig:case} (E) analyzes DGPR across different levels of rollout disagreement. We define examples with $D=0$ as the consensus group and divide the remaining examples into low ($0<D\leq0.075$), medium ($0.075<D\leq0.15$), and high ($D>0.15$) disagreement groups. We have the following observations. \textit{\textbf{Firstly,}} Disagreement provides an effective uncertainty signal. As $D$ increases from consensus to high disagreement, the anchor cell- and batch-level accuracies decrease from $89.9\%$ to $46.4\%$ and from $60.5\%$ to $5.2\%$ for Qwen3-8B. \textit{\textbf{Secondly,}} DGPR is most effective on medium disagreement samples. It improves cell- and batch-level accuracy by $0.29$ and $2.49$ for Qwen3-8B, and by $0.31$ and $0.81$ for Qwen3-4B. These samples contain meaningful alternative assignments while retaining a relatively reliable anchor, making them suitable for anchor--challenger reconsideration. \textit{\textbf{Thirdly,}} high disagreement reveals the limitation of pairwise refinement. DGPR improves cell-level accuracy but decreases batch-level accuracy by $0.27$ and $0.51$ for Qwen3-8B and Qwen3-4B. This indicates that local corrections do not necessarily recover the global matching, highlighting the coupling between cell identification and batch matching.


\subsection{Ablation Study}
To quantify the contribution of each component in \method{}, we conduct ablation studies with Qwen3-8B, as shown in Table~\ref{tab:ablation_study}. We compare five variants: (1) \method{} w/o CRFT, which removes the reasoning fine-tuning initialization and directly performs ROPSD from the base model; (2) \method{} w/o ROPSD, which removes the routed on-policy training stage; (3--4) two ROPSD variants that ablate its key designs, including \method{} w/o Dense Reward, which replaces the local--global dense reward with the sparse Cell-o1 reward, and \method{} w/o O1 Fallback, which disables offline OpenAI-O1 trajectory supervision when no qualified sibling is available; and (5) \method{} w/o DGPR, which uses single-pass inference. All ablations degrade average performance, with the largest drops caused by removing CRFT (6.01/12.86) and ROPSD (5.01/15.02), followed by the dense reward (3.63/8.25), DGPR (1.37/0.99), and O1 fallback (1.06/4.37). These results highlight the contribution of each component to the overall effectiveness of our proposed \method{}.


\subsection{Sensitivity Analysis}
We evaluate the sensitivity of \method{} to key hyperparameters (see Figure~\ref{fig:case} (B)). The first analysis studies the number of inference rollouts $M$, which determines the candidate reasoning trajectories used in DGPR. We observe that both accuracy generally improve as $M$ increases. However, the performance gain saturates when $M$ becomes large, while the computational cost continues to increase. Therefore, we set $M=8$ by default to achieve a trade-off between performance and efficiency. The second analysis studies the entropy weighting coefficient $\beta$, which controls the contribution of dynamic weighting. We observe that accuracy first improves and then decrease as $\beta$ increases, while the performance remains stable around $\beta=1$. We use $\beta=1$ as default, as it corresponds to the natural unscaled weighting and avoids dataset-specific tuning. The third analysis studies the consensus threshold $\gamma$, which controls when refinement is triggered. As $\gamma$ increases, both the trigger and acceptance rates increase. A lower threshold leaves correctable errors unrefined, whereas a higher threshold over-refines stable predictions; we therefore set $\gamma=0.75$ as the default. Overall, our \method{} demonstrates robust performance across various settings, highlighting its low sensitivity.



\subsection{Case Study}
We present a case study on an example from CellPuzzles to illustrate how \method{} handles coupled cell-type assignment errors (see Figure~\ref{fig:case} (A)). As shown in Figure~\ref{fig:case}(a), closely related kidney cell types require subtle marker distinctions, including \textit{SLC12A3} for DCT, \textit{UMOD} and \textit{SLC12A1} for TAL, and \textit{ALDOB}, \textit{FABP1}, and \textit{MIOX} for proximal tubule cells. Cell-o1 misinterprets these markers, causing local annotation errors that occupy incorrect labels and propagate through the one-to-one matching constraint. In contrast, \method{} uses rollout disagreement to localize uncertain cells, compares a medoid anchor with a supported challenger, and reconsiders the uncertain assignments while preserving stable predictions. By jointly considering canonical markers and the global matching constraint, \method{} correctly assigns Cells 10--12 while preserving the stable assignments of the remaining cells, achieving a correct batch-level assignment. This example demonstrates the ability of \method{} to resolve globally coupled annotation errors.

\section{Discussion}

In this study, we proposed \method{}, a routed on-policy self-distillation framework for LLM-based single-cell annotation. The core of \method{} is to exploit rollout \textbf{consensus} for privileged training supervision and rollout \textbf{disagreement} for uncertainty-aware refinement. \method{} first conducts reasoning fine-tuning, then routes rollouts to outcome-based reward optimization and privileged self-distillation. Finally, \method{} identifies uncertain predictions through disagreement and performs refinement. Experiments 
demonstrate that \method{} outperforms baselines in cell-level and batch-level annotation while achieving strong generalization and biological interpretability.


Further analyses provide evidence for the mechanisms underlying these improvements. Firstly, the adaptive routing dynamics show that successful on-policy siblings become the primary source of privileged supervision and that the optimization gradually shifts from self-distillation toward GRPO as the policy improves. Secondly, rollout disagreement is strongly associated with prediction errors, suggesting that it provides a reliable uncertainty signal. Thirdly, the kidney cell-type case study shows that disagreement identifies ambiguous assignments, while refinement uses canonical marker genes to correct them, demonstrating how uncertainty-aware refinement in \method{} mitigates error propagation in batch-level annotation.



Despite its promising performance, \method{} require multiple rollouts, increasing both training and inference costs. Future work could develop more efficient uncertainty estimators or adaptive sampling strategies that use fewer trajectories for confident examples. Moreover, \method{} could be extended by retrieving biological evidence from large-scale cell atlases and reference datasets~\citep{Quake2021TheTS,Lotfollahi2021MappingSD,yu-etal-2025-scrag}. These extensions may support cross-tissue or rare cell type annotation in real-world scenarios~\citep{10.1093/bioadv/vbad030}.

\section{Conflicts of interest}
The authors declare that they have no competing interests.

\section{Funding}
Zhiyin Yu, Songtao Huang, Shuaiyu Zhang and Shiyang Feng were supported by the China Postdoctoral Science Foundation (Grant No. 2025M781483).

\section{Data availability}
Code and data are available at
\url{https://github.com/YuZhiyin/scOPD}.

\section{Author contributions statement}

Z.Y. developed the methodology, conducted the experimental implementation, and prepared the manuscript. S.F., X.Y., S.H. and Y.M. contributed to scientific discussion and experimental design. W.Y., S.Z , X.W. and C.C. contributed to data curation and literature review. Z.W., B.Z., L.B. and X.L. supervised the project, provided critical feedback, and contributed to manuscript refinement.   

\section{Acknowledgments}
The authors thank the anonymous reviewers for their valuable suggestions. Zhiyin Yu, Songtao Huang, Shuaiyu Zhang and Shiyang Feng were supported by the China Postdoctoral Science Foundation (Grant No. 2025M781483).

\bibliographystyle{oup-abbrvnat}
\bibliography{reference}

\end{document}